\documentclass{bmvc2k}

\usepackage{amsfonts}
\usepackage{lipsum}
\usepackage{multicol, float, array}
\usepackage{multirow}
\usepackage{graphicx}

\title{Structured Consistency Loss for Semi-Supervised Semantic Segmentation}

\addauthor{JongMok Kim}{win98man1@gmail.com}{1}
\addauthor{JooYoung Jang}{jyjang1090@snu.ac.kr}{2}
\addauthor{HyunWoo Park}{hw.park@graphics.snu.ac.kr}{3}
\addauthor{SeongAh Jeong}{seongah@knu.ac.kr}{4}

\addinstitution{
SNUAILAB
}
\addinstitution{
SNUAILAB
}
\addinstitution{
Seoul National Univ.
}
\addinstitution{
KyeongBook Univ.
}

\runninghead{Student, Prof, Collaborator}{BMVC Author Guidelines}

\begin{document}

\maketitle

\begin{abstract}
The consistency loss has the potential for semi-supervised learning in segmentation and localization with various recent problems as well as classification.
The conventional methods based on consistency loss, however, are inefficiently applied to semantic segmentation task regarded as pixel-wise classification; the consistency loss is calculated only in matched pixel pairs and averaged out.
To this end, we propose a \textit{structured consistency loss} for semi-supervised semantic segmentation to promote the consistency by considering the inter-pixel correlation. 
Specifically, the proposed method dramatically improves the computational complexity in the collaboration with cutmix \citep{yun2019cutmix}.
It is shown via experiments that the Cityscapes benchmark results with validation data and test data are 81.9mIoU and 83.84mIoU respectively. 
This ranks the \textit{first} place on the pixel-level semantic labeling task of Cityscapes benchmark suite. 
To the best of our knowledge, this is the first study to achieve the state-of-the-art performance with semi-supervised learning approach in semantic segmentation which provides the insight of the applicability of the semi-supervised technique to the various types of learning-based systems.
\end{abstract}

%%%%%%%%%%%%%%%%%%%%%%%%%%%%%%%%%%%%%%%%%%%%%%%%%%%%%%%%%%%%%%%%%%%%%%%%%%%%%%%%%%%%%%%%%%%%%%%%%%%%%%INTRODUCTION%%%%%%%%%%%%%%%%%%%%%%%%%%%%%%%%%%%%%%%%%%%%%%%%%%%%%%%%
\section{Introduction}
With the deep learning approach of \citet{krizhevsky2012imagenet} introduced in ILSVRC-2012 \citep{ILSVRC15}, deep learning-based image processing has acquired tremendous progress. 
Especially, the supervised learning has been paid increasing attention since the learning performance of the network is greatly affected by both quality and quantity of the labeled data.  
Meanwhile, the supervised learning requires the manually labeled data, which is time-consuming and expensive.
When it comes to semantic segmentation, pixel-by-pixel annotation is required, which particularly leads to the exorbitant preparation of the labeled data. 
Recently, a various techniques such as active learning \citep{mackowiak2018cereals}, interactive segmentation \citep{maninis2018deep}, weakly-supervised leaning \citep{lee2019ficklenet} and so on have been developed to solve the labeling cost problem in semantic segmentation. 

Semi-supervised learning has been introduced to address the ever-increasing size of modern data combined with the difficulty of obtaining label.
In particular, semi-supervised learning improves the network performance with the use of relatively smaller number of labeled data, when a large number of unlabeled data is available.
A simple build up of unlabeled data can be easily collected from various data sources via web crawling, vehicle logging, and etc.
Especially, the manual labelling process to allow to make the collected unlabeled data useful in learning process is costly compared to other tasks in the progress of semantic segmentation. 
To address the cost issue of labeling, the semi-supervised learning technique best suits the semantic segmentation which requires expensive labeled data. 

Prior studies \citep{tarvainen2017mean, xie2019unsupervised} employ the semi-supervised learning techniques for image classification task in a various ways, and show the significant improvement in accuracy. 
However, the semantic segmentation is a different task compared to the classification in that while the prediction of classification results in a class vector, the semantic segmentation performs the structured prediction per regional location and predicts structural characteristics of regions.
Prior studies \citep{liu2019structured,xie2018improving} investigate the between-pixel relationship and verify that the structural relationship between pixels is important in semantic segmentation.

In this paper, we propose a semi-supervised learning with the structured consistency loss for semantic segmentation.
The proposed loss forces the predictions of segmentation network consistent in not only pixel-wise, but also inter-pixel relationship, which allows the network to learn more powerful generalization capabilities to predict in harmony with neighboring pixels.
Furthermore, when cutmix \citep{yun2019cutmix} is incoporated, network can get the better generalization performance with lowering GPU memory utilization by restricting the regions to calculate the structured consistency loss in cutmix box.
Via numerical results, it is shown that the proposed method ranks the first place with mIOU 83.84 in the Cityscapes benchmark pixel-level semantic labeling task \citep{Cordts2016Cityscapes}.
We note that this is the first study to verify that the semi-supervised learning can achieve the state-of-the-art performance in semantic segmentation using the Cityscapes benchmark. 
In addition, the proposed semi-supervised learning technique can be applied in parallel with other researches for further improvements since our contribution is not in network architecture, but in learning techniques.

%%%%%%%%%%%%%%%%%%%%%%%%%%%%%%%%%%%%%%%%%%%%%%%%%%%%%%%%%%%%%%%%%%%%%%%%%%%%%%%%%%%%%%%%%%%%%%%%%%%%%%Related Works%%%%%%%%%%%%%%%%%%%%%%%%%%%%%%%%%%%%%%%%%%%%%%%%%%%%%%%
\section{Related Work} \label{related_works}
\textbf{Semantic Segmentation}.\hspace{1mm} 
The early period of semantic segmentation approaches are mostly based on Fully Convolutional Network (FCN) \citep{sermanet2013overfeat, long2015fully}.
To improve the performance of earlier segmentation models, the loss of spatial information is mediated with the use of encoder-decoder architecture \citep{noh2015learning, ronneberger2015u, badrinarayanan2017segnet} or dilated convolution (a.k.a. Atrous Convolution) expanding the receptive field \citep{chen2014semanticdeeplabv1, yu2015multi}. 
For the further enhancement on the localization performance, \citep{chen2017deeplabv2} applies the Atrous Spatial Pyramid Pooling (ASPP) in semantic segmentation, and PSPNet \citep{zhao2017pyramid} proposes a feature pyramid pooling module to gather global contextual information around the image object or stuff. 
More recently, \citet{chen2018encoderdeeplabv3+} suggestes the well-organized architecture with combining encoder-decoder architecture and dilated convolution, which have been followed by many subsequent methods to achieve state-of-the-art performance \citep{takikawa2019gated, zhuang2018dense, li2019global}.
\citet{zhu2019improving} presents the inspiring improvement of results thanks to labelling of video image using temporal information, boundary relaxation loss to address the boundary issue and class uniform sampling for the class imbalance problem. \\
%To make our contribution clearly visible, we employed the DeepLabV3Plus \citep{chen2018encoderdeeplabv3+}, boundary relaxation loss and class uniform sampling except for video label generation \citep{zhu2019improving} as the baseline .\\

\noindent\textbf{Semi-Supervised Learning}.\hspace{1mm} In recent years, the semi-supervised method has become a one of the most prominent theme, but is limitedly employed to classification task in earlier times. 
The usage of loss function computed on unlabeled data encourages the model to enhance generalization ability to unseen data in same domain. 
\citet{grandvalet2005semi} propose the entropy minimization loss to verify that the decision boundary tends to lie on a low density region of class distribution.
The consistency loss is suggested in \citet{laine2016temporal} so as to encourage the model to produce the same output distribution when its inputs are perturbed.
The consistency regularization loss plays a breakthrough role in the follow-up semi-supervised researches.
To employ the consistency loss efficiently, exponential-moving-average (EMA) technique which builds the teacher network by accumulating weights of student network to generate the more accurate guessed label is invented by \citet{tarvainen2017mean}. 
MixMatch \citep{berthelot2019mixmatch} combines entropy minimization, consistency loss, and MixUp regularization \citep{zhang2017mixup} for the generalization performance improvement of the network. 
Additionally, the results of \citet{xie2019unsupervised} nearly match for the performance of models trained on the full sets of CIFAR-10 with only using 10 \% of this dataset, thanks to sophisticated augmentation method with realistic noise. \\

\noindent\textbf{Structured Prediction}. \hspace{1mm}
The prediction results of semantic segmentation have an equal shape of input image, and also the prediction vector of each pixel has a strong correlation with each other, especially close one.
In \citet{xie2018improving}, a local pair-wise (8-neighbors) distillation is used to make an efficient feature distillation.
\citet{liu2019structured} presents an impressive progress of feature distillation by using pair-wise distillation, where the total inter-pixel similarities are calculated in a specific feature map, and the feature of teacher is distilled to student.

The knowledge distillation method which is similar to both the purpose and implementation of the consistency loss, is used to train the student network by forcing to resemble with the teacher network.
Accordingly, the distance functions are usually utilized as a loss function by both the knowledge distillation and consistency loss.
The consistency loss with cutmix \citep{yun2019cutmix} for semantic segmentation has been introduced in \citet{french2019consistency}. 
The authors in \citep{french2019consistency} investigate the difference between classification and semantic segmentation in terms of the low density region.
The low density region lies in the boundary of class distribution in classification, while it lies in a regional boundary of object in semantic segmentation.
They select the cutmix method instead of mixup to conserve the local boundarys of image, but with a conventional consistency loss.
%In this paper, we employ the structured knowledge distillation technique \citep{liu2019structured} and cutmix \citep{yun2019cutmix} in order to give the different perturbed images to teacher and student, while maintaining the structural information.

%%%%%%%%%%%%%%%%%%%%%%%%%%%%%%%%%%%%%%%%%%%%%%%%%%%%%%%%%%%%%%%%%%%%%%%%%%%%%%%%%%%%%%%%%%%%%%%%%%%%%%Methodology%%%%%%%%%%%%%%%%%%%%%%%%%%%%%%%%%%%%%%%%%%%%%%%%%%%%%%%%
\section{Methodology}
Since semantic segmentation can be regarded as a pixel by pixel classification, it is reasonable to simply apply the conventional consistency loss which is useful for the classification to the semantic segmentation.
The recent study \citep{french2019consistency} has made improvements by applying the consistency loss to the semantic segmentation. 
However, different from classification, the semantic segmentation has a characteristic of structured prediction that the predictions of pixels have correlation with each other. 
Therefore, if the existing consistency loss is used as it is, it is difficult to achieve the high performance improvement without consideration on the characteristic of semantic segmentation.
In order to boost the performance of semi-supervised semantic segmentation, in this section we introduce the structured consistency loss, encouraging the network to predict properly by focusing on inter-pixel relationship in cutmix box.

\subsection{Overall Training}
Overall training of our network is composed of labeled, and unlabeled image in a batch simultaneously.  The total loss $\mathcal{L}_{tot}$ is therefore a semi-supervised loss given as
\begin{equation}
\mathcal{L}_{tot} = \mathcal{L}_{x} + \mathcal{L}_{u},
\label{eq1}
\end{equation}

\begin{multicols}{2}
%\hspace{1cm}
\begin{figure}[H]
\centering
\includegraphics[width=1.0\linewidth, height = 1.0\linewidth]{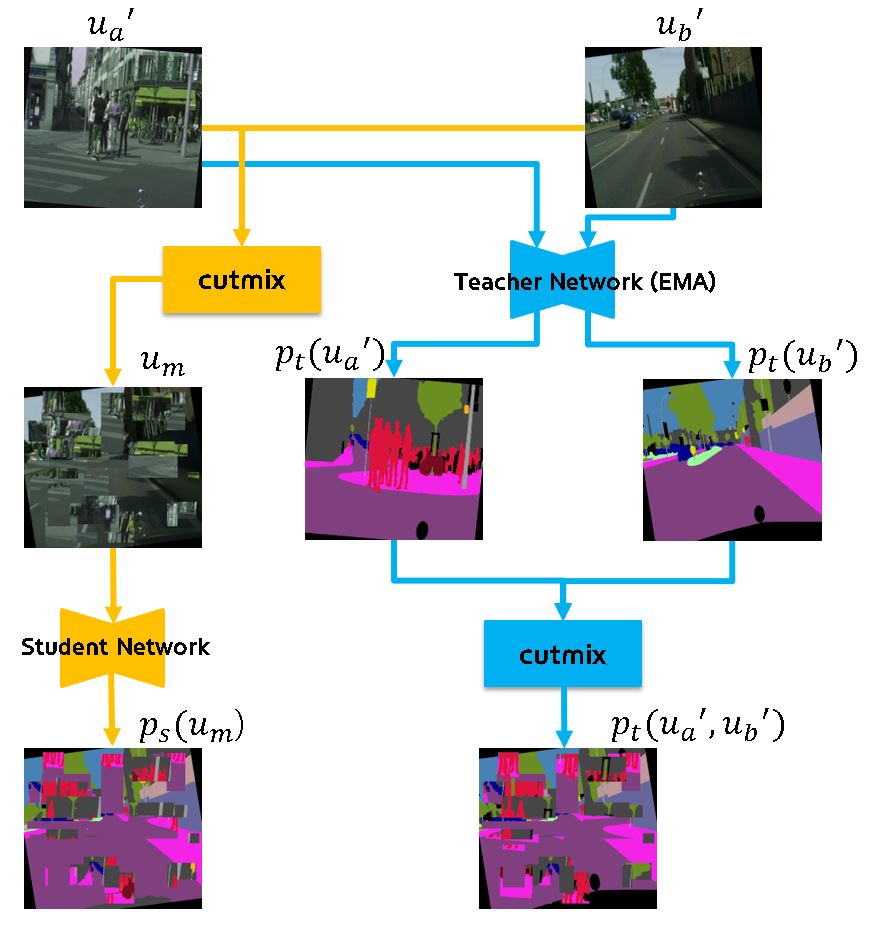}
\vspace{-0.5cm}
\caption{Architecture of the unsupervised training is depicted. Two predictions ($p_s(u_m)$, $p_{t}(\acute{u_a},\acute{u_b})$)  from the student network and teacher network are used to calculate the consistency loss and structured consistency loss.}
\label{Figure1}
\end{figure}

\begin{figure}[H]
\includegraphics[width=1.0\linewidth]{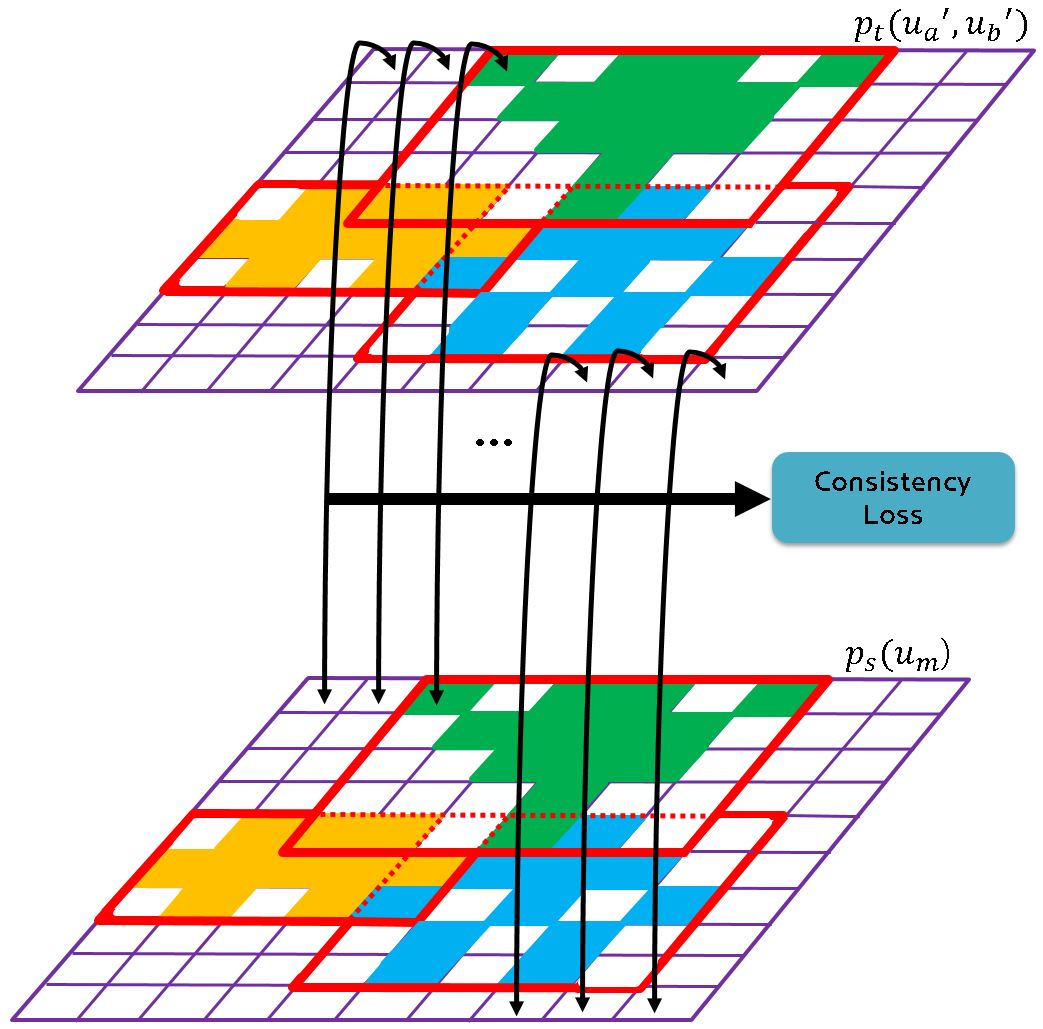}
\vspace{-0.5cm}
\caption{The calculation of the consistency loss is described. The consistency loss is calculated in each paired pixel and averaged out. For the simplicity, Image size is reduced to 10 x 10 and has three cutmix boxes whose boundary is red line.}
\label{Figure2}
\end{figure}
\end{multicols}

\noindent where $\mathcal{L}_{u}$ is a loss with unlabeled image $u$ and $\mathcal{L}_{x}$ is a supervised cross entropy loss with boundary label relaxation \citep{zhu2019improving} for semantic segmentation with labeled image $x$ and written as
\begin{equation}
\mathcal{L}_x = -log\sum\limits_{C\in \mathcal{N}_w}P(C).
\label{eq2}
\end{equation}
$P(\cdot)$ and $\mathcal{N}_w$ are softmax probability for each class, and the set of classes within a $w$ by $w$ window boundary region, respectively.
Note that $\mathcal{L}_x$ reduces to the standard cross entropy loss with one-hot label when $w$ equals one, i.e., $w$=1.

\subsection{Architecture for unsupervised training}
The cutmix augmentation is originally designed for classification task, where it mixes samples by cutting box and pasting from one sample into another. 
Following \citet{french2019consistency}, we use the cutmix augmentation method for semantic segmentation by generating $N$ boxes with random size and position. 
The total area of boxes is approximately the half of the image dimension to make balance between two images $u_a$ and $u_b$.

Our main architecture of unsupervised approach is described in Fig.\ref{Figure1}.
First, we generate randomly augmented input images ($\acute{u_a}$, $\acute{u_b}$) from the original images ($u_a$, $u_b$) by random augmentation used in \citet{zhu2019improving} except the cutmix, and create a cutmix image ($u_m$) from the augmented images.
The teacher network makes prediction results ($p_t(\acute {u_a}) $, $p_t(\acute{u_b}) $) from the input images ($\acute{u_a}$, $\acute{u_b}$) and cutmix the results ($p_{t}(\acute{u_{a}},\acute{u_{b}})$). The student network uses the cutmix results ($p_{t}(\acute{u_{a}},\acute{u_{b}})$) as a \textit{guessed label} introduced by \citet{berthelot2019mixmatch}.
By getting the cutmix image ($u_m$) as input, the student network makes prediction ($p_s(u_m)$). 
Finally, we calculate two consistency losses using the prediction of student network and teacher network.
Note that teacher and student network have the same network architecture, but the weights of teacher network are not from its own weights using gradient update, but from the student network by EMA manipulation.  
The guessed label can promote the generalization performance to the student network since the label is made by two original images information when the input to the student network is a complex image as several patches stuck on.
The unlabeled loss term is separated into two parts such as consistency loss and structured consistency loss. 
The formula of unlabeled loss is given as follows:
\begin{equation}
\mathcal{L}_{u} = \lambda_c\mathcal{L}_{c} + \lambda_{sc}\mathcal{L}_{sc},
\label{eq3}
\end{equation}
\vspace{-0.5cm}
\begin{equation}
\label{eq4}
\mathcal{L}_{c} = \frac{1}{H \times W} \sum\limits_{i \in \mathbb{T}} ||p_s(u_{m,i})-p_{t}(\acute{u_{a,i}},\acute{u_{b,i}})||^2,
\end{equation}
where $\lambda_{c}$ and  $\lambda_{sc}$ are hyperparameters for each loss term of $\mathcal{L}_{c}$ and $\mathcal{L}_{sc}$, respectively, and $\mathcal{L}_{sc}$ is the structured consistency loss which would be addressed in more detail in the following subsection.
$\mathcal{L}_{c}$ is the conventional consistency loss as an average of pixel-wise squared L2 loss described in Fig.\ref{Figure2} and Eq.\ref{eq4}. 
$p_s(u_{m,i})$ represents a prediction of student network for the \textit{i}th pixel of cutmix unlabeled image $u_m$, while $p_{t}(\acute{u_{a,i}},\acute{u_{b,i}})$ represents a cutmix of predictions of the teacher network for the \textit{i}th pixel for unlabeled images $\acute{u_a}$ and $\acute{u_b}$.
We denote a set of all the pixel indices in the image as $\mathbb{T}=\{1,2,...,H\times W\}$,

\subsection{Structured consistency loss}
As explained in section \ref{related_works}, we utilize a concept of pair-wise knowledge distillation technique suggested by \citet{liu2019structured} for the structured consistency loss. 
Although the knowledge distillation and the consistency loss differs in underlying philosophy, the concept of consideration on inter-pixel relationship of pair-wise knowledge distillation is the great help in extending the conventional consistency loss into the structural one.
The proposed structured consistency loss can be written as
\begin{equation}
a_{ij} = \mathbf{p}_i^{T}\mathbf{p}_j/(||\mathbf{p}_i||||\mathbf{p}_j||),
\label{eq5}
\end{equation}
\vspace{-0.5cm}
\begin{equation}
\mathcal{L}_{sc} = \frac{1}{(H \times W)^2}\sum\limits_{i\in\mathbb{T}}\sum\limits_{j\in\mathbb{T}}||a_{ij}^s - a_{ij}^t||^2 ,
\label{eq6}\\
\end{equation}
where $a_{ij}^{t}$ and $a_{ij}^{s}$ denote the similarity between the \textit{i}th pixel and the \textit{j}th pixel produced from the teacher network and student network, respectively, and $\mathbf{p}_i$ represents the prediction vector of the \textit{i}th pixel.
%In the previous work \citep{liu2019structured}, the prediction vector ($\mathbf{p} $) in Eq.\ref{eq5} is calculated at the feature map rather than the prediction map due to feature distillation effects.
%On the contrary, the structured consistency loss have to be calculated at the image domain, because its goal is to get the same prediction from the perturbed image.
%To this end, there is the problem to overcome to calculate the structured consistency loss in image domain, that is, the calculation of structured loss consumes a lot of GPU memory usage since the complexity is $(H \times W)^2$ which is not feasible in the image domain with the same size as the input.
However, the structured consistency loss derived from Eq.\ref{eq6} is not efficient because the most of pixel pairs are far from each other, have very low correlation, and therefore have little effect on performance.
% However, the structured consistency loss derived from Eq.\ref{eq6} is not efficient because the most of pixel pairs have little correlation since they are too far from each other. Those pairs has little effect on performance.
Moreover, Eq.\ref{eq6} requires extremely high computation cost for computing all of pixel pairs in prediction map.

\begin{figure}[t]
\begin{center}
%\framebox[4.0in]{$\;$} 
%\includegraphics[width=0.8\textwidth,natwidth=610,natheight=642]{Figure2_3}
\includegraphics[width=1.0\linewidth]{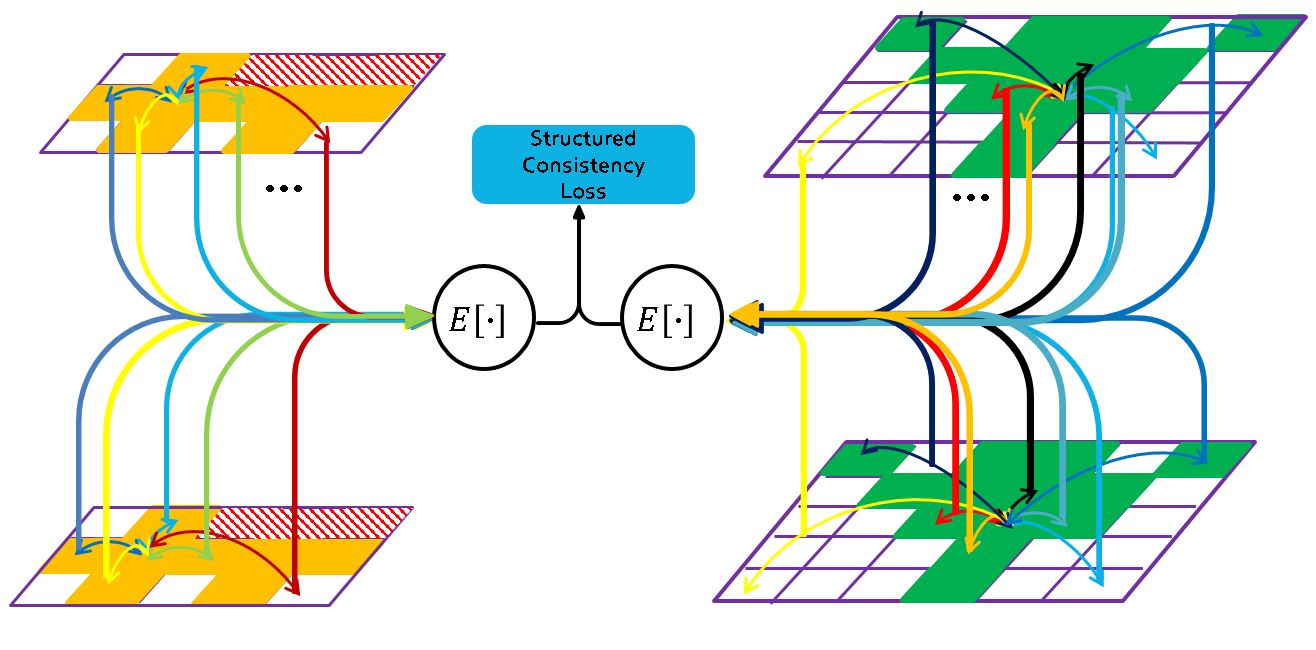}
\end{center}
\vspace{-0.3cm}
\caption{The calculation of structured consistency loss is described. Assume that the above boxes are derived from teacher network, while the below boxes are derived from student network. Inter-pixel cosine similarity is calculated at each cutmix box. The hatched area of left cutmix box which is covered by the other box is not included in calculation. For simplicity, we set $N_{box}$=2.} 
\vspace{-0.3cm}
\label{Figure3}
\end{figure}
In order to handle the above problems, we employ the cutmix augmentation, which limits the pixel pairs inside the local region and reduces computing complexity dramatically.
By calculating the inter-pixel cosine similarity within $N$ cutmix box separately, we can reduce the computing complexity approximately $N$ times.
The detailed operation is described in Fig.\ref{Figure3}.
Moreover, by using cutmix, the network can learn the ability to make the accurate predictions with limited local information, which can provide the prediction similar to that of global area.
The structured consistency loss with cutmix is represented as
\begin{equation}
a_{n,ij} = \mathbf{p}_{n,i}^{T}\mathbf{p}_{n,j}/(||\mathbf{p}_{n,i}||||\mathbf{p}_{n,j}||),
\label{eq7}
\end{equation}
\vspace{-0.5cm}
\begin{equation}
\mathcal{L}_{sc} = \frac{1}{N}\sum\limits_{n=1}^{N}\frac{1}{|\mathbb{T}_n|^2}\sum\limits_{i\in\mathbb{T}_n}\sum\limits_{j\in\mathbb{T}_n}||a_{n,ij}^s - a_{n,ij}^t||^2,
\label{equ8}
\end{equation}
where $a_{n,ij}$ represents the similarity between the \textit{i}th pixel and the \textit{j}th pixel in \textit{n}th cutmix box, $\mathcal{L}_{sc}$ is the structured consistency loss which is the sum of $N$ cutmix boxes calculation, $N$ is the number of cutmix boxes, and $\mathbb{T}_n = \{1,2,3,...,H_n \times W_n\}$ denotes a set of all the pixel indices in the \textit{n}th cutmix box except for the region covered by other cutmix box.

To make the structured consistency loss more efficient, we restrict the number of cutmix boxes to $N_{box}$ and the maximum number of pairs to be calculated in each box to $N_{pair}$. 
In Fig.\ref{Figure3}, we eliminate the region covered by another cutmix box, and also randomly drop out the similarity vectors excessing over $N_{pair}$.
Then, we derive the efficient and effective structured consistency loss as follows:

\begin{equation}
\mathbb{T}_{n,drop}^{s} = Drop\boldsymbol{\langle}\{(i,j)|\forall i\in \mathbb{T}_n, \forall j\in \mathbb{T}_n \}, N_{pair}\boldsymbol{\rangle},
\label{equ9}
\end{equation}
\vspace{-0.75cm}
\begin{equation}
\mathcal{L}_{sc} = \frac{1}{N_{box}}\sum\limits_{n=N-N_{box}+1}^{N}\frac{1}{|\mathbb{T}_{n,drop}^{s}|}\sum\limits_{(i,j)\in\mathbb{T}_{n,drop}^{s}}||a_{n,ij}^s - a_{n,ij}^t||^2,
\label{equ10}
\end{equation}
where the customized function $Drop\boldsymbol{\langle} \mathbb{X}, n \boldsymbol{\rangle}$ randomly removes the elements of $\mathbb{X}$ to keep the maximum number of elements as $n$.
In the process of attaching the cutmix box, the box attached earlier is likely to be covered by the box attached behind, and then the structed consistency loss is calculated using posterior $ N_{box}  $ of boxes.
In addition, the loss is not calculated via masking the area covered by another cutmix box. 
By doing this, only the pixels that are structurally-relevant within the cutmix box can learn the relationship each other.

%%%%%%%%%%%%%%%%%%%%%%%%%%%%%%%%%%%%%%%%%%%%%%%%%%%%%%%%%%%%%%%%%%%%%%%%%%%%%%%%%%%%%%%%%%%%%%%%%%%%%%Experimentss%%%%%%%%%%%%%%%%%%%%%%%%%%%%%%%%%%%%%%%%%%%%%%%%%%%%%%%

\section{Experiments}
\subsection{Implementation details}\label{exp_imp}

\textbf{Network Structures}.\hspace{1mm} We adopt a state-of-the-art segmentation architecture, Deeplabv3+ with WideResNet38 from \citet{zhu2019improving}. 
% \textit{Output stride} of encoder is set to 8 and that of the low level feature transferred to decoder is set to 2.
\textit{Output stride} of the last layer of encoder and the low level feature transferred to decoder is 8, and 2 respectively.
% The teacher network to generate the guessed label is EMA model \citep{tarvainen2017mean}.
EMA \citep{tarvainen2017mean} is used for the teacher network to generate the guessed label.
Inference with the test sets and the validation sets is also carried out by this model. 
The parameter value of EMA weight is set to 0.999 which is averaged out in every training steps by following \citep{tarvainen2017mean}. \\

\noindent\textbf{Training Procedure}.\hspace{1mm}
We use a SGD optimizer with the polynomial learning rate policy. 
Following the setup of \citet{zhu2019improving}, the initial learning rate is 0.002, the power set to 1.0, weight decay of 0.001 and momentum of 0.9.
We use the synchronized batch normalization with a batch size of 2 per GPU, one for labeled data and the other for unlabeled one, on 8 V100 GPUs. 
The training epoch is set to 175 based on the labeled data, while, for unlabeled data, we adopt cutmix augmentation. 
From the empirical experiments, we set the hyper-parameters $N$, $N_{box}$, $N_{pair}$ $\lambda_{c}$ and $\lambda_{sc}$ to 32, 16, 9000, 20 and 3, respectively. 
We use additional training skills such as Mapillary pre-training, class uniform sampling and boundary label relaxation \citep{zhu2019improving} so as to get the strong baseline network.\\

\noindent\textbf{Cityscapes}. \hspace{1mm}
Cityscapes is a widely accepted validation medium among recent studies \citep{zhu2019improving, french2019consistency} owing to its high-quality labeled image database.
Cityscapes offers the pixel-level annotations of 5,000 images and coarse annotations of 20K images. 
The reliable pixel-level annotations are also split into three subsets, that is, training, validation and test, each of which consists of 2,975, 500 and 1,525 images, respectively, and gauge the performance of our methodology. 
We also use coarsely annotated images for class uniform sampling to overcome the imbalance between classes, which are also applied with ignoring the annotations associated with unlabeled images in unlabeled training phase for greater learning exposure.

\begin{table}[ht]
\fontsize{5.2}{6}\selectfont
\centering
\caption{Comparison of Per-class mIoU results of recent methods on Cityscapes with the proposed method marked as "Ours" and "Ours+". 
Including both published and unpublished methods, the results obtained by proposed method provide the best overall performance. Note that the results for "Ours" is obtained from training data only while "Ours+" is obtained from training and validation data.}
\vspace{0.2cm}
\label{test_cityscape}
\hspace{-0.1cm}
\begin{tabular}{l||c|c|c|c|c|c|c|c|c|c|c|c|c|c|c|c|c|c|c||c}
\hline
\hspace{-0.182cm} Method \hspace{-0.182cm} & \hspace{-0.182cm} road \hspace{-0.182cm} & \hspace{-0.182cm} s.walk \hspace{-0.182cm} & \hspace{-0.182cm} build.\hspace{-0.182cm}  & \hspace{-0.182cm} wall \hspace{-0.182cm} & \hspace{-0.182cm} fence \hspace{-0.182cm} & \hspace{-0.182cm} pole \hspace{-0.182cm} & \hspace{-0.182cm} t-light \hspace{-0.182cm} & \hspace{-0.182cm} t-sign \hspace{-0.182cm} & \hspace{-0.182cm} veg \hspace{-0.182cm} & \hspace{-0.182cm} terrain \hspace{-0.182cm} & \hspace{-0.182cm} sky \hspace{-0.182cm} & \hspace{-0.182cm} person \hspace{-0.182cm} & \hspace{-0.182cm} rider \hspace{-0.182cm} & \hspace{-0.182cm} car \hspace{-0.182cm} & \hspace{-0.182cm} truck \hspace{-0.182cm} & \hspace{-0.182cm} bus \hspace{-0.182cm} & \hspace{-0.182cm} train \hspace{-0.182cm} & \hspace{-0.182cm} motor \hspace{-0.182cm} & \hspace{-0.182cm} bike \hspace{-0.182cm} & \hspace{-0.182cm} mIoU \hspace{-0.182cm} \\ \hline \hline 

\hspace{-0.182cm} PSPNet \citep{zhao2017pyramid} \hspace{-0.182cm} & \hspace{-0.182cm} 98.7 \hspace{-0.182cm} & \hspace{-0.182cm} 86.9 \hspace{-0.182cm} & \hspace{-0.182cm} 93.5 \hspace{-0.182cm} & \hspace{-0.182cm} 58.4 \hspace{-0.182cm} & \hspace{-0.182cm} 63.7 \hspace{-0.182cm} & \hspace{-0.182cm} 67.7 \hspace{-0.182cm} & \hspace{-0.182cm} 76.1 \hspace{-0.182cm} & \hspace{-0.182cm} 80.5 \hspace{-0.182cm} & \hspace{-0.182cm} 93.6 \hspace{-0.182cm} & \hspace{-0.182cm} 72.2 \hspace{-0.182cm} & \hspace{-0.182cm} 95.3 \hspace{-0.182cm} & \hspace{-0.182cm} 86.8 \hspace{-0.182cm} & \hspace{-0.182cm} 71.9 \hspace{-0.182cm} & \hspace{-0.182cm} 96.2 \hspace{-0.182cm} & \hspace{-0.182cm} 77.7 \hspace{-0.182cm} & \hspace{-0.182cm} 91.5 \hspace{-0.182cm} & \hspace{-0.182cm} 83.6 \hspace{-0.182cm} & \hspace{-0.182cm} 70.8 \hspace{-0.182cm} & \hspace{-0.182cm} 77.5 \hspace{-0.182cm} & \hspace{-0.182cm} 81.2 \hspace{-0.182cm} \\ \hline

\hspace{-0.182cm} DeeplabV3+ \citep{chen2018encoderdeeplabv3+} \hspace{-0.182cm} & \hspace{-0.182cm} 98.7 \hspace{-0.182cm} & \hspace{-0.182cm} 87.0 \hspace{-0.182cm} & \hspace{-0.182cm} 93.9 \hspace{-0.182cm} & \hspace{-0.182cm} 59.4 \hspace{-0.182cm} & \hspace{-0.182cm} 63.7 \hspace{-0.182cm} & \hspace{-0.182cm} 71.4 \hspace{-0.182cm} & \hspace{-0.182cm} 78.2 \hspace{-0.182cm} & \hspace{-0.182cm} 82.2 \hspace{-0.182cm} & \hspace{-0.182cm} 94.0 \hspace{-0.182cm} & \hspace{-0.182cm} 73.0 \hspace{-0.182cm} & \hspace{-0.182cm} 95.8 \hspace{-0.182cm} & \hspace{-0.182cm} 88.0 \hspace{-0.182cm} & \hspace{-0.182cm} 73.0 \hspace{-0.182cm} & \hspace{-0.182cm} 96.4 \hspace{-0.182cm} & \hspace{-0.182cm} 78.0 \hspace{-0.182cm} & \hspace{-0.182cm} 90.9 \hspace{-0.182cm} & \hspace{-0.182cm} 83.9 \hspace{-0.182cm} & \hspace{-0.182cm} 73.8 \hspace{-0.182cm} & \hspace{-0.182cm} 78.9 \hspace{-0.182cm} & \hspace{-0.182cm} 82.1 \hspace{-0.182cm} \\ \hline

\hspace{-0.182cm} Gated-SCNN \citep{takikawa2019gated} \hspace{-0.182cm} & \hspace{-0.182cm} 98.7 \hspace{-0.182cm} & \hspace{-0.182cm} 87.4 \hspace{-0.182cm} & \hspace{-0.182cm} 94.2 \hspace{-0.182cm} & \hspace{-0.182cm} 61.9 \hspace{-0.182cm} & \hspace{-0.182cm} 64.6 \hspace{-0.182cm} & \hspace{-0.182cm} 72.9 \hspace{-0.182cm} & \hspace{-0.182cm} \textbf{79.6} \hspace{-0.182cm} & \hspace{-0.182cm} 82.5 \hspace{-0.182cm} & \hspace{-0.182cm} \textbf{94.3} \hspace{-0.182cm} & \hspace{-0.182cm} 73.3 \hspace{-0.182cm} & \hspace{-0.182cm} \textbf{96.2} \hspace{-0.182cm} & \hspace{-0.182cm} 88.3 \hspace{-0.182cm} & \hspace{-0.182cm} 74.2 \hspace{-0.182cm} & \hspace{-0.182cm} \textbf{96.6} \hspace{-0.182cm} & \hspace{-0.182cm} 77.2 \hspace{-0.182cm} & \hspace{-0.182cm} 90.2 \hspace{-0.182cm} & \hspace{-0.182cm} 87.7 \hspace{-0.182cm} & \hspace{-0.182cm} 72.6 \hspace{-0.182cm} & \hspace{-0.182cm} 79.4 \hspace{-0.182cm} & \hspace{-0.182cm} 82.8 \hspace{-0.182cm} \\ \hline

\hspace{-0.182cm} DRN-CRL \citep{zhuang2018dense}  \hspace{-0.182cm} & \hspace{-0.182cm} 98.8 \hspace{-0.182cm} & \hspace{-0.182cm} 87.7 \hspace{-0.182cm} & \hspace{-0.182cm} 94.0 \hspace{-0.182cm} & \hspace{-0.182cm} \textbf{65.0} \hspace{-0.182cm} & \hspace{-0.182cm} 64.2 \hspace{-0.182cm} & \hspace{-0.182cm} 70.2 \hspace{-0.182cm} & \hspace{-0.182cm} 77.4 \hspace{-0.182cm} & \hspace{-0.182cm} 81.6 \hspace{-0.182cm} & \hspace{-0.182cm} 93.9 \hspace{-0.182cm} & \hspace{-0.182cm} 73.5 \hspace{-0.182cm} & \hspace{-0.182cm} 95.8 \hspace{-0.182cm} & \hspace{-0.182cm} 88.0 \hspace{-0.182cm} & \hspace{-0.182cm} 74.9 \hspace{-0.182cm} & \hspace{-0.182cm} 96.5 \hspace{-0.182cm} & \hspace{-0.182cm} \textbf{80.8} \hspace{-0.182cm} & \hspace{-0.182cm} 92.1 \hspace{-0.182cm} & \hspace{-0.182cm} 88.5 \hspace{-0.182cm} & \hspace{-0.182cm} 72.1 \hspace{-0.182cm} & \hspace{-0.182cm} 78.8 \hspace{-0.182cm} & \hspace{-0.182cm} 82.8 \hspace{-0.182cm} \\ \hline

\hspace{-0.182cm} GALD-Net \citep{li2019global} \hspace{-0.182cm} & \hspace{-0.182cm} 98.8 \hspace{-0.182cm} & \hspace{-0.182cm} 87.7 \hspace{-0.182cm} & \hspace{-0.182cm} 94.2 \hspace{-0.182cm} & \hspace{-0.182cm} 65.0 \hspace{-0.182cm} & \hspace{-0.182cm} \textbf{66.7} \hspace{-0.182cm} & \hspace{-0.182cm} \textbf{73.1} \hspace{-0.182cm} & \hspace{-0.182cm} 79.3 \hspace{-0.182cm} & \hspace{-0.182cm} 82.4 \hspace{-0.182cm} & \hspace{-0.182cm} 94.2 \hspace{-0.182cm} & \hspace{-0.182cm} 72.9 \hspace{-0.182cm} & \hspace{-0.182cm} 96.0 \hspace{-0.182cm} & \hspace{-0.182cm} \textbf{88.4} \hspace{-0.182cm} & \hspace{-0.182cm} \textbf{76.2} \hspace{-0.182cm} & \hspace{-0.182cm} 96.5 \hspace{-0.182cm} & \hspace{-0.182cm} 79.8 \hspace{-0.182cm} & \hspace{-0.182cm} 89.6 \hspace{-0.182cm} & \hspace{-0.182cm} 87.7 \hspace{-0.182cm} & \hspace{-0.182cm} \textbf{74.1} \hspace{-0.182cm} & \hspace{-0.182cm} \textbf{79.9} \hspace{-0.182cm} & \hspace{-0.182cm} 83.3 \hspace{-0.182cm} \\ \hline

\hspace{-0.182cm} Video Propagation \citep{zhu2019improving} \hspace{-0.182cm}  & \hspace{-0.182cm} 98.8 \hspace{-0.182cm} & \hspace{-0.182cm} 87.8 \hspace{-0.182cm} & \hspace{-0.182cm} 94.2 \hspace{-0.182cm} & \hspace{-0.182cm} 64.0 \hspace{-0.182cm} & \hspace{-0.182cm} 65.0 \hspace{-0.182cm} & \hspace{-0.182cm} 72.4 \hspace{-0.182cm} & \hspace{-0.182cm} 79.0 \hspace{-0.182cm} & \hspace{-0.182cm} 82.8 \hspace{-0.182cm} & \hspace{-0.182cm} 94.2 \hspace{-0.182cm} & \hspace{-0.182cm} 74.0 \hspace{-0.182cm} & \hspace{-0.182cm} 96.1 \hspace{-0.182cm} & \hspace{-0.182cm} 88.2 \hspace{-0.182cm} & \hspace{-0.182cm} 75.4 \hspace{-0.182cm} & \hspace{-0.182cm} 96.5 \hspace{-0.182cm} & \hspace{-0.182cm} 78.8 \hspace{-0.182cm} & \hspace{-0.182cm} \textbf{94.0} \hspace{-0.182cm} & \hspace{-0.182cm} \textbf{91.6} \hspace{-0.182cm} & \hspace{-0.182cm} 73.7 \hspace{-0.182cm} & \hspace{-0.182cm} 79.0 \hspace{-0.182cm} & \hspace{-0.182cm} 83.5 \hspace{-0.182cm} \\ \hline

\hspace{-0.182cm} Ours \hspace{-0.182cm}  & \hspace{-0.182cm} \textbf{98.8} \hspace{-0.182cm} & \hspace{-0.182cm} \textbf{88.2} \hspace{-0.182cm} & \hspace{-0.182cm} \textbf{94.3} \hspace{-0.182cm} & \hspace{-0.182cm} 64.5 \hspace{-0.182cm} & \hspace{-0.182cm} 65.3 \hspace{-0.182cm} & \hspace{-0.182cm} 72.6 \hspace{-0.182cm} & \hspace{-0.182cm} 79.3 \hspace{-0.182cm} & \hspace{-0.182cm} \textbf{82.9} \hspace{-0.182cm} & \hspace{-0.182cm} 94.2 \hspace{-0.182cm} & \hspace{-0.182cm} \textbf{74.2} \hspace{-0.182cm} & \hspace{-0.182cm} 96.1 \hspace{-0.182cm} & \hspace{-0.182cm} 88.4 \hspace{-0.182cm} & \hspace{-0.182cm} 75.6 \hspace{-0.182cm} & \hspace{-0.182cm} 96.6 \hspace{-0.182cm} & \hspace{-0.182cm} 79.5 \hspace{-0.182cm} & \hspace{-0.182cm} 93.7 \hspace{-0.182cm} & \hspace{-0.182cm} 91.2 \hspace{-0.182cm} & \hspace{-0.182cm} 73.7 \hspace{-0.182cm} & \hspace{-0.182cm} 79.4 \hspace{-0.182cm} & \hspace{-0.182cm} \textbf{83.6} \hspace{-0.182cm} \\ \hline\hline

\hspace{-0.182cm} Tencent AI Lab \hspace{-0.182cm}  & \hspace{-0.182cm} 98.6 \hspace{-0.182cm} & \hspace{-0.182cm} 86.9 \hspace{-0.182cm} & \hspace{-0.182cm} 94.1 \hspace{-0.182cm} & \hspace{-0.182cm} 63.5 \hspace{-0.182cm} & \hspace{-0.182cm} 63.0 \hspace{-0.182cm} & \hspace{-0.182cm} 70.7 \hspace{-0.182cm} & \hspace{-0.182cm} 77.7 \hspace{-0.182cm} & \hspace{-0.182cm} 80.2 \hspace{-0.182cm} & \hspace{-0.182cm} 94.0 \hspace{-0.182cm} & \hspace{-0.182cm} 73.1 \hspace{-0.182cm} & \hspace{-0.182cm} 95.9 \hspace{-0.182cm} & \hspace{-0.182cm} 87.8 \hspace{-0.182cm} & \hspace{-0.182cm} 74.5 \hspace{-0.182cm} & \hspace{-0.182cm} 96.3 \hspace{-0.182cm} & \hspace{-0.182cm} 82.8 \hspace{-0.182cm} & \hspace{-0.182cm} 94.3 \hspace{-0.182cm} & \hspace{-0.182cm} 90.4 \hspace{-0.182cm} & \hspace{-0.182cm} 74.0 \hspace{-0.182cm} & \hspace{-0.182cm} 77.5 \hspace{-0.182cm} & \hspace{-0.182cm} 82.9 \hspace{-0.182cm} \\ \hline

\hspace{-0.182cm} GGCF  \hspace{-0.182cm}  & \hspace{-0.182cm} 98.8 \hspace{-0.182cm} & \hspace{-0.182cm} 87.8 \hspace{-0.182cm} & \hspace{-0.182cm} 94.1 \hspace{-0.182cm} & \hspace{-0.182cm} 66.0 \hspace{-0.182cm} & \hspace{-0.182cm} 66.1 \hspace{-0.182cm} & \hspace{-0.182cm} 71.1 \hspace{-0.182cm} & \hspace{-0.182cm} 78.4 \hspace{-0.182cm} & \hspace{-0.182cm} 82.2 \hspace{-0.182cm} & \hspace{-0.182cm} 94.1 \hspace{-0.182cm} & \hspace{-0.182cm} \textbf{74.5} \hspace{-0.182cm} & \hspace{-0.182cm} 95.8 \hspace{-0.182cm} & \hspace{-0.182cm} 88.0 \hspace{-0.182cm} & \hspace{-0.182cm} 74.0 \hspace{-0.182cm} & \hspace{-0.182cm} 96.5 \hspace{-0.182cm} & \hspace{-0.182cm} 79.9 \hspace{-0.182cm} & \hspace{-0.182cm} 92.4 \hspace{-0.182cm} & \hspace{-0.182cm} 90.8 \hspace{-0.182cm} & \hspace{-0.182cm} 71.8 \hspace{-0.182cm} & \hspace{-0.182cm} 78.4 \hspace{-0.182cm} & \hspace{-0.182cm} 83.2 \hspace{-0.182cm} \\ \hline

\hspace{-0.182cm} iFLYTEK-CV \hspace{-0.182cm}  & \hspace{-0.182cm} 98.7 \hspace{-0.182cm} & \hspace{-0.182cm} 87.1 \hspace{-0.182cm} & \hspace{-0.182cm} 94.1 \hspace{-0.182cm} & \hspace{-0.182cm} 64.4 \hspace{-0.182cm} & \hspace{-0.182cm} 65.4 \hspace{-0.182cm} & \hspace{-0.182cm} 71.2 \hspace{-0.182cm} & \hspace{-0.182cm} 77.9 \hspace{-0.182cm} & \hspace{-0.182cm} 82.2 \hspace{-0.182cm} & \hspace{-0.182cm} 94.0 \hspace{-0.182cm} & \hspace{-0.182cm} 73.5 \hspace{-0.182cm} & \hspace{-0.182cm} 96.0 \hspace{-0.182cm} & \hspace{-0.182cm} 88.3 \hspace{-0.182cm} & \hspace{-0.182cm} 75.7 \hspace{-0.182cm} & \hspace{-0.182cm} 96.5 \hspace{-0.182cm} & \hspace{-0.182cm} \textbf{83.3} \hspace{-0.182cm} & \hspace{-0.182cm} \textbf{94.7} \hspace{-0.182cm} & \hspace{-0.182cm} \textbf{92.4} \hspace{-0.182cm} & \hspace{-0.182cm} 74.3 \hspace{-0.182cm} & \hspace{-0.182cm} 79.0 \hspace{-0.182cm} & \hspace{-0.182cm} 83.6 \hspace{-0.182cm} \\ \hline

\hspace{-0.182cm} openseg-group \hspace{-0.182cm} & \hspace{-0.182cm} 98.8 \hspace{-0.182cm} & \hspace{-0.182cm} 88.3 \hspace{-0.182cm} & \hspace{-0.182cm} 94.3 \hspace{-0.182cm} & \hspace{-0.182cm} \textbf{66.9} \hspace{-0.182cm} & \hspace{-0.182cm} \textbf{66.7} \hspace{-0.182cm} & \hspace{-0.182cm} \textbf{73.3} \hspace{-0.182cm} & \hspace{-0.182cm} \textbf{80.2} \hspace{-0.182cm} & \hspace{-0.182cm} \textbf{83.0} \hspace{-0.182cm} & \hspace{-0.182cm} 94.2 \hspace{-0.182cm} & \hspace{-0.182cm} 74.1 \hspace{-0.182cm} & \hspace{-0.182cm} 96.0 \hspace{-0.182cm} & \hspace{-0.182cm} 88.5 \hspace{-0.182cm} & \hspace{-0.182cm} 75.8 \hspace{-0.182cm} & \hspace{-0.182cm} 96.5 \hspace{-0.182cm} & \hspace{-0.182cm} 78.5 \hspace{-0.182cm} & \hspace{-0.182cm} 91.8 \hspace{-0.182cm} & \hspace{-0.182cm} 90.2 \hspace{-0.182cm} & \hspace{-0.182cm} 73.4 \hspace{-0.182cm} & \hspace{-0.182cm} 79.3 \hspace{-0.182cm} & \hspace{-0.182cm} 83.7 \hspace{-0.182cm} \\ \hline

\hspace{-0.182cm} Ours+ \hspace{-0.182cm} & \hspace{-0.182cm} \textbf{98.9} \hspace{-0.182cm} & \hspace{-0.182cm} \textbf{88.4} \hspace{-0.182cm} & \hspace{-0.182cm} \textbf{94.3} \hspace{-0.182cm} & \hspace{-0.182cm} 65.2 \hspace{-0.182cm} & \hspace{-0.182cm} 65.9 \hspace{-0.182cm} & \hspace{-0.182cm} 72.8 \hspace{-0.182cm} & \hspace{-0.182cm} 79.5 \hspace{-0.182cm} & \hspace{-0.182cm} 83.0 \hspace{-0.182cm} & \hspace{-0.182cm} \textbf{94.3} \hspace{-0.182cm} & \hspace{-0.182cm} 74.3 \hspace{-0.182cm} & \hspace{-0.182cm} \textbf{96.1} \hspace{-0.182cm} & \hspace{-0.182cm} \textbf{88.6} \hspace{-0.182cm} & \hspace{-0.182cm} \textbf{75.9} \hspace{-0.182cm} & \hspace{-0.182cm} \textbf{96.6} \hspace{-0.182cm} & \hspace{-0.182cm} 79.3 \hspace{-0.182cm} & \hspace{-0.182cm} 93.8 \hspace{-0.182cm} & \hspace{-0.182cm} 91.5 \hspace{-0.182cm} & \hspace{-0.182cm} \textbf{74.8} \hspace{-0.182cm} & \hspace{-0.182cm} \textbf{79.7} \hspace{-0.182cm} & \hspace{-0.182cm} \textbf{83.8} \hspace{-0.182cm} \\ \hline

\end{tabular}
\end{table}

\subsection{Results}

\begin{figure}[h]
\begin{center}
\includegraphics[width=1.0\linewidth]{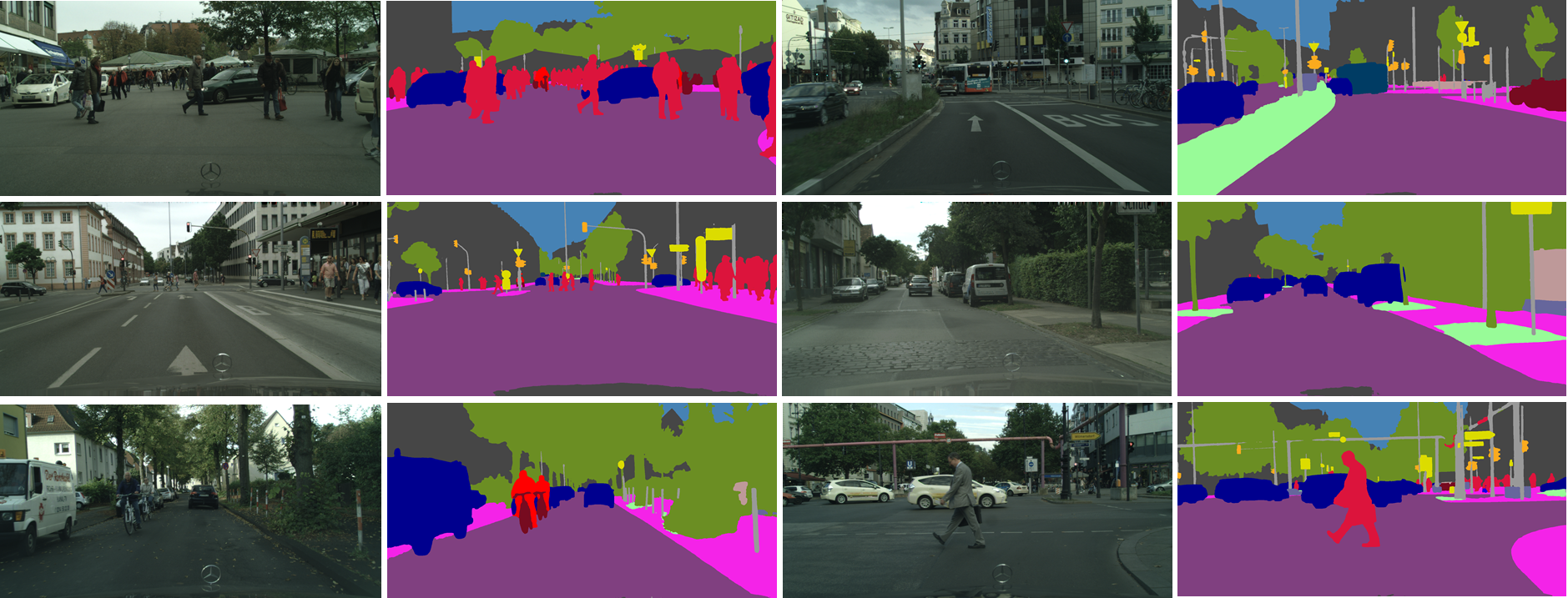}
\end{center}
\vspace{-0.5cm}
\caption{Qualitative results of our proposed method on the Cityscapes test set, where the pictures depict the predicted segmentation masks. }
\label{test_inference}
\end{figure}

The experimental results from our semi-supervised learning technique with the structured consistency loss are summarized in Table \ref{test_cityscape}.
The experiments are conducted on Cityscapes test images with multi-scale strategy (0.5, 1.0  and 2.0), horizontal-flip, and overlapping-tile methods following \citep{zhu2019improving}.
The performance results achieved by our proposed method (marked as "Ours" and "Ours+") rank the first place in the Cityscapes benchmark, including the results of unpublished studies.
It is also noticeable that the result with only training set exhibits the highest performance result compared with the performance results of published methods. 
Graphical results on the Cityscapes test set are presented in Fig.\ref{test_inference}. 

We believe that the our observed performance is attributed to (i) the effectiveness of structured consistency loss added to semi-supervised learning in semantic segmentation and (ii) the usage of coarsely labeled images in unlabeled training phase.
Extant studies of semantic segmentation use the coarsely labeled data only in the early training stage, or sporadically use to aid the segmentation of image with rare classes.
From our numerical results, the coarsely labeled data can be fully exploited not only by using the coarsely label in restricted way, but also by semi-supervised learning approach.

\subsection{Ablation Study}

\begin{figure}[h]
\begin{center}
\includegraphics[width=1.0\linewidth]{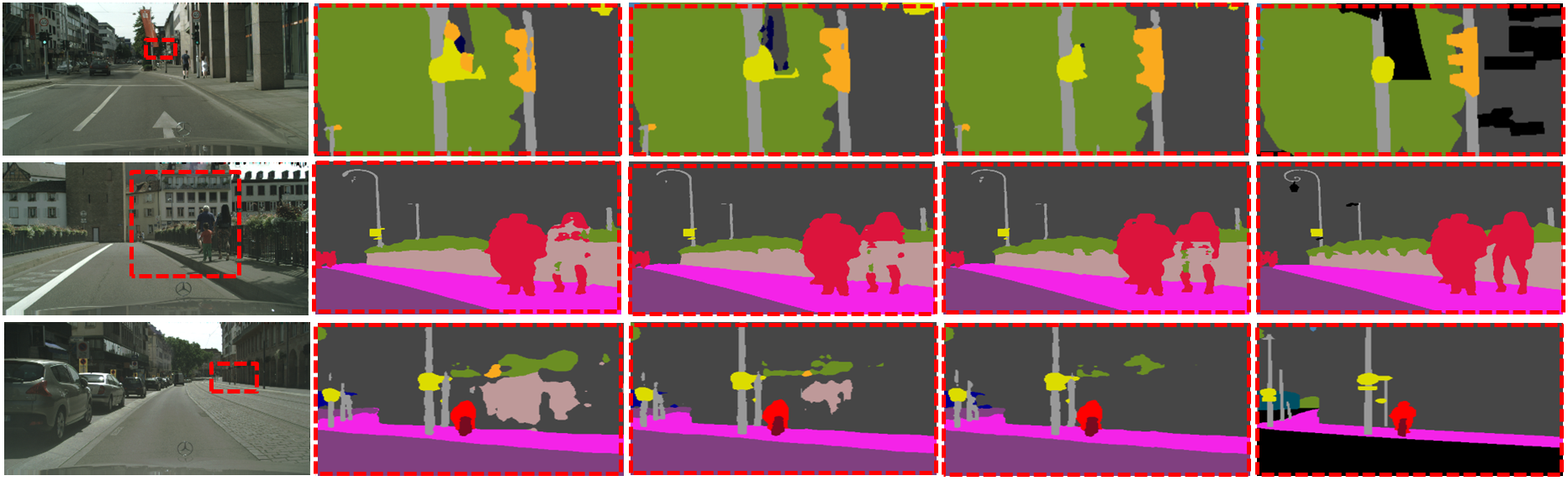}
\end{center}
\vspace{-0.5cm}
\caption{Qualitative results on the validation data of Cityscapes. From the left to the right are original images, supervised-only, the addition of consistency loss, the addition of structured consistency loss, and ground truth. Red box is zoomed-region where is the visually confusing.}
\label{val_compare}
\end{figure}

\noindent\textbf{Effect of Semi-Supervised Losses}.\hspace{1mm}
In this section, we conduct the additional experiments to demonstrate the superiority and effectiveness of two consistency losses.
We first train the baseline network with only the supervised method whose result is summarized in the fifth row of Table \ref{ablation_baseline}. 
% By adding consistency loss to the baseline, the mIoU increases to 81.55\% from 81.18\% by showing the practical implication of the semi-supervised learning with the cutmix in semantic segmentation. 
%  By adding consistency loss to the baseline, the mIoU increases to 81.55\% from 81.18\% showing that semi-supervised sementic segmentation has practical advantage with cutmix.
It is shown in Table \ref{ablation_baseline} that semi-supervised learning is beneficial to the semantic segmentation because the performance increases 0.37\% by adding consistency loss to the baseline, and performance further increases with the addition of the structured consistency loss to 81.90\%, which verifies the significant improvement by our proposed method. 
The qualitative results have been visually provided in Fig.\ref{val_compare} by allowing us to gauge the qualitative performance. 
Thanks to the inter-pixel consistency processed with the structured consistency loss, the proposed method exhibits the less errors in small and visually confusing region(s).\\

\noindent\textbf{Exponential-Moving-Average (EMA) Application}. \hspace{1mm}
The EMA is a well-known method for its great generalization ability \citep{tarvainen2017mean,berthelot2019mixmatch}, beneficial to apply for both building a teacher network and employing in validation (or test) phase. 
In recent studies in semi-supervised learning, however, not all the studies apply EMA to the weight of teacher network, or the inference network which implies that there is no rule of thumb.
%but use the same weight as that of student network and use the network of EMA weight only when validation, i.e., there is no rule of thumb.
In order to examine the performance dependency on EMA, we conduct an experiment whose results are summarized in Table \ref{ablation_ema}.
When the EMA weight is used in the teacher network, mIoU increases 0.135\% on average, while the use of EMA weight in validation seems to have no significant effect on performance. 
Since the use of EMA weight in teacher network and validation yields the greatest mIoU, we proceed to use the EMA weight in both.  

\begin{multicols}{2}
\begin{table}[H]
\centering
\fontsize{7}{7.5}\selectfont
\caption{Comparison of mIoU results of recent methods, supervised-only (baseline), consistency loss, and structured consistency loss on Cityscapes validation images.}
\vspace{0.35cm}
\label{ablation_baseline}
\begin{tabular}{ m{3cm}|c|c}
\hline
Method & mIoU (\%) & Gain (\%) \\ \hline \hline
DeepLabV3+ \citep{chen2018encoderdeeplabv3+} & 79.6 & \hspace{1.4mm}- \\ \hline
PSPNet \citep{zhao2017pyramid} & 79.7 & \hspace{1.4mm}- \\ \hline
Gated-SCNN \citep{takikawa2019gated} & 80.8 & \hspace{1.4mm}- \\ \hline
Video Propagation \citep{zhu2019improving} & 81.4 & \hspace{1.4mm}- \\ \hline\hline

Baseline & 81.17 & \hspace{1.4mm}0.0 \\ \hline
+ Consistency loss & 81.55 & +0.38 \\ \hline
+ Structured Consistency loss  & 81.90 & +0.35 \\ \hline

\end{tabular}
\end{table}

\begin{table}[H]
\centering
\small
\caption{Comparison of mIoU depending on whether exponential-moving-average (EMA) is applied to both/either teacher network and/or validation on Cityscapes image.}
\label{ablation_ema}
\vspace{0.35cm}
\begin{tabular}{c|c|c}
\hline
\multicolumn{2}{c|}{EMA} & \multirow{2}{*}{mIoU (\%)} \\ \cline{1-2}
Teacher      & Validation     &                  \\ \hline\hline
$\mathsf{X}$            & $\mathsf{X}$             & 81.77                             \\ \hline
$\mathsf{X}$            & $\mathsf{O}$             & 81.74                             \\ \hline
$\mathsf{O}$            & $\mathsf{X}$             & 81.88                             \\ \hline
$\mathsf{O}$            & $\mathsf{O}$             & 81.90                            \\ 
\hline
\end{tabular}
\end{table}
\end{multicols}
\vspace{-0.2cm}

%%%%%%%%%%%%%%%%%%%%%%%%%%%%%%%%%%%%%%%%%%%%%%%%%%%%%%%%%%%%%%%%%%%%%%%%%%%%%%%%%%%%%%%%%%%%%%%%%%%%%%Conclusion%%%%%%%%%%%%%%%%%%%%%%%%%%%%%%%%%%%%%%%%%%%%%%%%%%%%%%%
\section{Conclusions}
In this paper, we propose the structured consistency loss in semi-supervised learning for semantic segmentation. With the cutmix augmentation, the structured consistency loss fully exploits the relationship among the local regions, and enhances the network generalization, which can also be simultaneously applied to the general network.
It is verified via numerical results that our proposed method achieves the state-of-the-art premier performance in the Cityscapes benchmark suite, not only ranking the first place only among publication results, but also among all results including unpublished results. 
Additionally, it is verified that the semi-supervised learning is highly effective to solve practical real world problems under data-insufficiency when it is accompanied with cutmix augmentation as well as the structured consistency loss. 
%The proposed method still can be further improved by the random sampling with $N_{pair}$ and $N_{box}$ which we leave as a future work.

%%%%%%%%%%%%%%%%%%%%%%%%%%%%%%%%%%%%%%%%%%%%%%%%%%%%%%%%%%%%%%%%%%%%%%%%%%%%%%%%%%%%%%%%%%%%%%%%%%%%%%%%%%%%%%%%%%%%%%%%%%%%%%%%%%%%%%%%%%%%%%%%%%%%%%%%%%%%%%%%%%%%%%%%%%%%%%%%%%%%%%%%%%%%%%%%%%%%%%%%%%%%%%%%%%

\bibliography{bib}
\end{document}